# Recursive Self-Improvement for Long-Horizon Role-Playing Dialogue

Y. L. Li, Y. H. Wu

Henan Chengxi Technology Co., Ltd.

## Abstract

Long-horizon role-playing dialogue requires consistent character identity, reliable memory, and context-sensitive interaction. We present a bounded recursive self-improvement (RSI) framework that revises original-character memory, stable persona instructions, and a policy for constructing turn-level prompts. Repair outcomes also inform revisions to the updater's diagnostic rules, edit routing, and candidate selection. We specify the updater as a versioned policy with explicit inputs, editable fields, acceptance gates, and stopping conditions. An internal evaluation covers 50 original characters and five models, with two 100-round conversations per character–model pair: 500 conversations per condition. Across the pooled evaluated replies, the bad-response rate decreases from 14.00% (6,973/49,824) to 4.94% (2,459/49,793), a 9.06-percentage-point reduction and a 64.71% relative reduction. In 600 blinded human comparisons, RSI receives 294 wins, 187 ties, and 119 losses, yielding a 71.19% win rate among decisive judgments. The results support improved aggregate dialogue quality for the complete framework on the internal test distribution.

## 1 Introduction

Role-playing conversational agents must maintain a character across a developing interaction. A locally fluent reply can still reverse a relationship, forget an established event, repeat an earlier plot beat, or decide an action on the interlocutor's behalf. These failures become difficult to diagnose when the same reply depends on persistent character instructions, retrieved memories, and a dynamically assembled prompt.

Public role-play work provides useful foundations for this problem. RoleLLM studies role-conditioned instruction tuning [2], while CharacterEval and EmoCharacter examine character portrayal and emotional fidelity [3,4]. MiniMax's account of Talkie/Xingye motivates multi-turn evaluation based on detectable dialogue errors [1]. These approaches motivate a system that can identify concrete failures and revise the responsible character component.

Our framework makes three components editable: character memory, stable persona instructions, and the policy that chooses turn-level context and guidance. An updater diagnoses failures and proposes localized repairs. Repair history then informs changes to the updater itself, allowing subsequent iterations to use different diagnostic rules, edit choices, and candidate priorities. The foundation model and immutable character canon remain outside this update interface.

We evaluate the complete system against the existing baseline using AI-generated internal test conversations and blinded human comparisons. The observed bad-response reduction is 64.71% relative to the baseline. The method section makes the update mechanism explicit, including how repair evidence reaches the next updater version and how validation feedback is separated from final testing. The empirical comparison measures the combined system effect.

## 2 Related Work

RoleLLM organizes character profiles and role-conditioned training, while CharacterEval supplies a multidimensional Chinese role-play benchmark [2,3]. EmoCharacter emphasizes emotional fidelity [4]. These studies motivate evaluating character behavior beyond surface fluency. Memory-Driven Role-Playing examines the use of persona knowledge, and PersonaForge studies personality consistency [5,6]. Generative Agents provides a memory, reflection, retrieval, and planning architecture for coherent agent behavior [9].

RMTBench evaluates role play through user motivations, and PersonaArena considers dynamic social interactions [7,8]. MiniMax's Role-Play Bench uses simulated multi-turn interactions and an error taxonomy organized around Worlds, Stories, and User Preferences [1].

Self-Refine and Reflexion use feedback to improve outputs or subsequent attempts [10,11]. Our framework additionally represents the diagnostic and repair policy as an editable object. This makes it possible to distinguish a changed character configuration from a changed procedure for producing future repairs. Recursively generated data can also lose distributional coverage [12], motivating scenario diversity and explicit protection of character-specific behavior.

## 3 System Setting

The baseline contains an OC profile, memory storage and retrieval, stable character instructions, and a contextual prompt assembled before each response. For character $c$ at iteration $k$, the editable configuration is

$$B_{c,k} = (M_{c,k}, P_{c,k}, \pi_{c,k}).$$

$M$ contains attributed events and relationship state; $P$ defines durable behavioral instructions; and $\pi$ selects memories and response guidance from the current dialogue state. The immutable canon $C_c$ records identity, core traits, world rules, and relationship boundaries. Canon facts take precedence over inferred memories or preferences.

Each memory is scoped to its character and dialogue instance. A corrected event in one simulated story must not become a fact in another story. Reusable changes concern memory handling or prompting rules; episode-specific content remains local. Baseline and RSI comparisons use the same base model within each model pairing. Model-weight optimization is outside the RSI mechanism described here.

## 4 RSI Improvement Cycle

One iteration begins with a character configuration $B_{c,k}$ and an updater $I_k$. The system generates test dialogue, diagnoses failures (Step 1), proposes character edits (Step 2), and validates those edits (Step 3). It then uses the repair outcomes to propose an updater revision (Step 4). An accepted updater governs diagnosis and edit selection in the next iteration. Figure 1 shows these two connected feedback paths.

Sections 4.1 and 4.2 define the data boundaries and editable state. Sections 4.3 and 4.4 describe diagnosis and repair generation. Section 4.5 defines character-update validation and selection. Section 4.6 describes updater revision and validation. Section 4.7 traces one example through the complete cycle.

### 4.1 Cycle Boundaries and Data Access

Transient adjustments within a dialogue can change the next response's context. Durable changes are submitted at iteration boundaries as versioned patches. A temporary response correction therefore does not automatically rewrite an OC's stable instructions or another dialogue's memories.

The protocol assigns optimization scenarios $D_k$ to diagnosis and candidate generation. Validation scenarios $V_k$ select character and updater versions. A final test set $H$ is reserved for evaluating a frozen configuration and cannot be accessed by the proposer, updater, or checkpoint selector. Scenario families are grouped before partitioning, including shared story templates and close paraphrases. This grouping makes a split more meaningful than withholding a differently worded version of a familiar scene.

Validation feedback is part of the optimization process. Even a pass/fail summary can guide later choices and produce adaptation to a repeatedly queried validation set. Each run therefore records its validation queries, candidate budget, and scenario-batch identifiers. Fresh validation batches are assigned to later cycles under a fixed schedule. If a batch is reused, its feedback is treated as selection data; independent performance claims come from the untouched final test set.

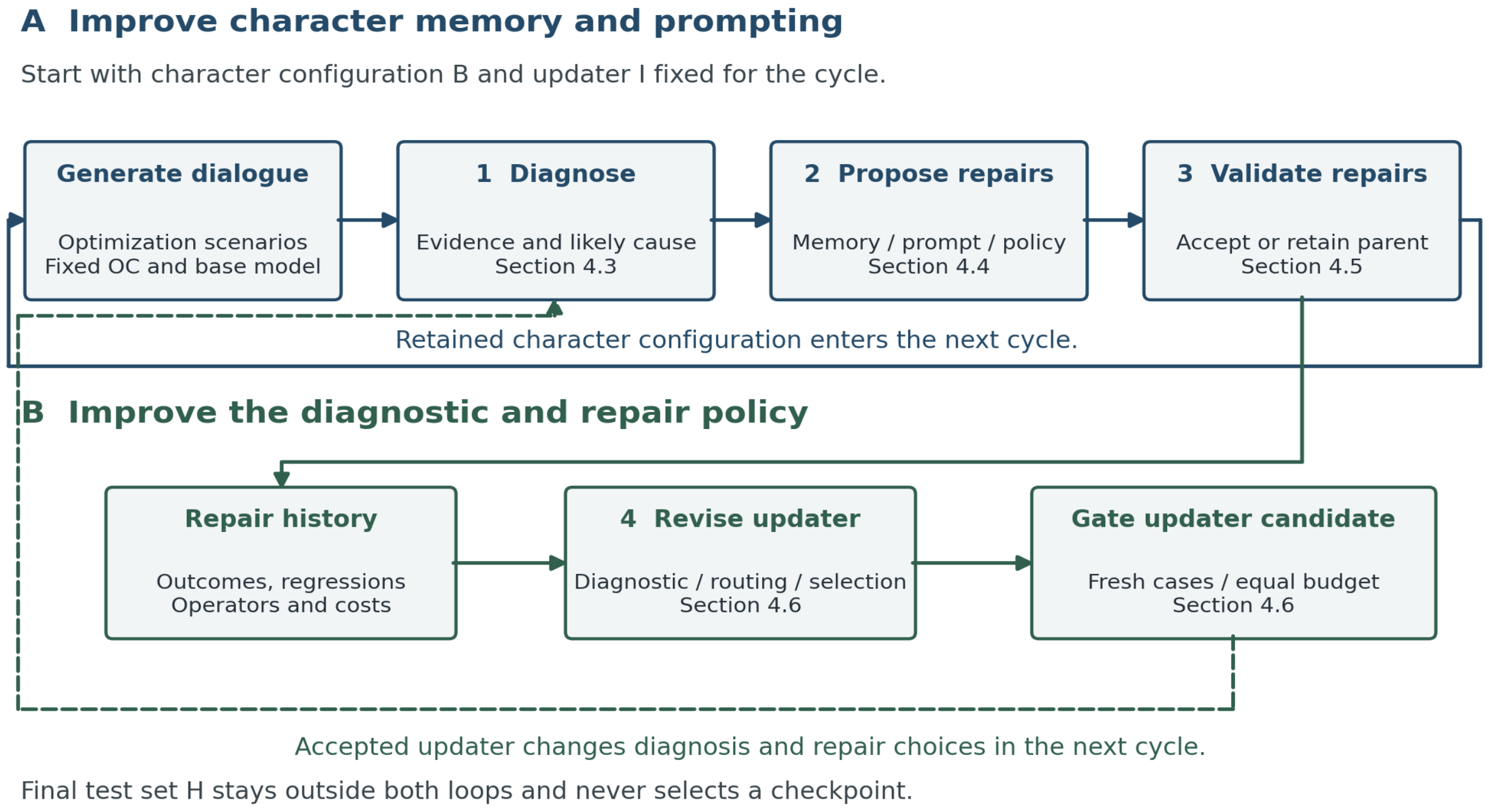

Figure 1. The upper loop repairs character memory and prompting. The lower loop revises the updater using repair outcomes and feeds the accepted policy into the next cycle. Validation and final-test access are controlled outside the updater.

Model-version, simulator, and rubric configurations are frozen before final evaluation.

The updater may inspect optimization traces and structured repair outcomes. It cannot rewrite the canonical OC specification, scoring rubric, acceptance thresholds, data partitions, or final-test access policy. These permissions are enforced by the orchestration layer, which also records rejected proposals and retained parent versions.

### 4.2 Editable State and Updater Representation

The character interface separates content from instructions. A memory item stores an event or relation, source dialogue span, story time, character and episode identifiers, and revision status. Stable instructions $P$ express voice, motivations, and interaction rules. The dynamic policy $\pi$ maps the current scene to selected memories, emotional stance, pacing guidance, and an invitation for the interlocutor's next action. Updating $\pi$ changes the rule used to construct future prompts, rather than one prompt instance.

We represent the updater as a structured policy

$$I_k = (d_k, r_k, s_k).$$

Here, $d_k$, $r_k$, and $s_k$ denote diagnostic, routing, and selection rules, respectively. The diagnostic rules define error conditions, required evidence, and when to abstain. Routing rules map a diagnosis and its context to an edit surface and permitted operator. Selection rules rank candidate repairs and allocate a fixed evaluation budget. Rules are serialized as versioned records and rendered into the updater's instructions; they are distinct from the OC's own persona prompt.

**Table 1. Updater fields and their permitted changes.**

| Field | Stored content | Example of an allowed revision |
|---|---|---|
| Diagnostic rules | Error condition, evidence requirements, abstention rule | Require a prior event citation before labeling a contradiction |
| Routing rules | Context predicate, target component, edit operator | Route an unretrieved event to retrieval repair rather than persona editing |
| Selection rules | Candidate priorities, tie breaks, budget allocation | Prioritize small repairs with fewer observed regressions |

Each revision contains a parent version, a minimal rule diff, supporting repair identifiers, its intended effect, and a rollback target. The updater cannot expand its own permissions. Threshold values, candidate limits, and stopping budgets are run-level configuration rather than self-editable rules.

### 4.3 Step 1 Diagnose Dialogue Failures

Step 1 takes a optimization trajectory, the OC canon, and the current memory and prompt configuration. It outputs a failure record containing the affected turn, error category, supporting span, violated fact or interaction requirement, and a likely responsible component. The diagnostic rules in $I_k$ determine which evidence is sufficient and when the diagnosis should remain unresolved.

For example, an OC may overlook a previous disagreement because the event was never stored, because retrieval missed it, or because the response ignored correctly retrieved evidence. These cases require different repairs. Inspecting the memory record and the assembled prompt helps distinguish them. A vague judgment that a reply is unengaging is insufficient to authorize a durable change unless it can be connected to the scene objective or an explicit comparison criterion.

The diagnostic rubric includes world consistency, narrative behavior, and interaction failures. Multiple valid reactions may exist for the same scene. The detector therefore identifies an evidenced violation or records uncertainty; it does not require every OC to produce a common ideal answer. Failure records pass to Step 2 with their provenance intact.

### 4.4 Step 2 Propose Character Edits

Step 2 holds $I_k$ fixed while its routing and selection rules generate a bounded candidate set. Each candidate targets a specific part of $M$, $P$, or $\pi$ and states an expected effect. A missing event may require a source-backed memory correction. Recurrent voice drift may call for a stable instruction change. Premature intimacy or stagnant pacing may call for a contextual rule in the dynamic policy.

The candidate set includes a no-change control. Candidates are evaluated against the triggering scenario and contrasting scenes where the previous behavior was appropriate. This checks whether a repair merely suppresses a legitimate character behavior everywhere. Each patch is reversible and records the character, episode scope, edited field, supporting failure, and protected invariants.

A patch generator may propose several alternatives, but the configured candidate budget bounds the search. Selection uses the current updater's priorities; it cannot increase its own model-call allowance. An uncertain diagnosis can lead to additional optimization evidence or abstention instead of an unsupported memory or personality rewrite.

### 4.5 Step 3 Validate and Select Character Updates

**Step 3 accepts or rejects character changes.** A candidate is produced by applying the current updater to optimization evidence:

$$\widetilde{B}_{c,k+1} = U_{I_k}(B_{c,k}, D_k).$$

Validation compares the candidate with its parent under the same scene specification and scoring rules. A valid candidate must improve its target outcome by the configured minimum and satisfy the protected constraints. For a bad-response target, improvement is the parent's bad-response rate minus the candidate's rate. Canon violations or unauthorized actions on behalf of the interlocutor can trigger hard rejection. Other permitted tolerances are set before candidate evaluation.

After selection, the orchestrator retains an accepted $B_{c,k+1}$ or keeps $B_{c,k}$. Both successful and rejected repairs enter history $\mathcal{R}_k$. A history item links the diagnosis, chosen surface and operator, candidate diff, validation outcome, regression tags, and evaluation cost. The record preserves which examples supported diagnosis and which outcomes were used only for selection.

### 4.6 Step 4 Update and Validate the Improvement Policy

A fixed meta-update instruction asks a language model to propose a minimal change to the structured policy. Its inputs are $I_k$, representative successful and failed repair records, the editable-field schema, and the fixed resource and permission constraints. The instruction requires a cited repair pattern, a proposed rule diff, an expected downstream effect, and a counterexample where the rule should not apply. The resulting candidate policy is submitted to the updater gate:

$$\tilde{I}_{k+1} = A(I_k, \mathcal{R}_k).$$

The transition $A$ has four operations. First, group repair records by error context, target surface, and operator. Second, identify recurring mismatches, such as accepted local fixes that cause regressions elsewhere or repeated selection of an ineffective surface. Third, propose an atomic change to a diagnostic, routing, or selection rule. Fourth, apply the schema and permission checks before executing the candidate. Where evidence conflicts, the candidate can narrow a rule's conditions or add abstention rather than expanding its scope.

The meta-update instruction and validator are held fixed, while the diagnostic and repair policy evolves across iterations.

**Updater validation and selection.** The candidate updater and its parent receive the same fresh validation failures, starting character configurations, and candidate-generation budget. They each produce character repairs. The fixed gate compares the downstream effects of those repairs, rejecting updater revisions that violate protected constraints. Among eligible improvements, selection prioritizes validated repair gain; ties favor lower evaluation cost and smaller rule changes. Acceptance thresholds are configured outside the updater. The selected version becomes $I_{k+1}$, or the parent remains active.

Character-patch validation and updater validation use separate scenario batches. Updater validation may still influence later optimization through recorded outcomes; it is not the final performance test. The loop stops when the fixed iteration or call budget is exhausted, or when the configured stopping criterion finds no accepted updater improvement. The final checkpoint is selected from optimization and validation records before $H$ is evaluated. Repeatedly selecting checkpoints using $H$ would turn it into selection data and require a new final test set.

### 4.7 A Complete Iteration

Consider an illustrative OC that repeatedly advances intimacy before the interlocutor reciprocates. Step 1 identifies the pacing failure and cites the relationship state. Step 2 proposes a conditional dynamic-policy edit: maintain the OC's affectionate voice, but wait for reciprocal evidence before escalating the relationship. Step 3 checks whether that repair reduces the target failure while preserving appropriate affection in scenes that already contain reciprocal signals.

Suppose the repair history shows that similar dynamic-policy edits work while broad stable-prompt edits suppress the OC's voice. Step 4 can narrow the routing rule from “relationship error → stable prompt” to “premature escalation with intact persona evidence → dynamic pacing policy.” The updater candidate must then produce better repairs on fresh validation cases before acceptance. In the next cycle, the accepted routing rule changes which repair is tried first.

One iteration follows this sequence: freeze parent versions; generate and diagnose optimization dialogues; propose and validate character patches; record repair outcomes; propose and validate an updater revision; then continue with the retained versions. All decisions retain parent links for rollback and audit.

## 5 Experimental Setup

### 5.1 Test Scale and Comparison Conditions

The internal evaluation comprises 50 OCs and five base models. Each character–model pair is tested in two conversations of 100 rounds. Thus each OC has 10 conversations across the five models, and each condition has 500 conversations in total. A round contains an AI interlocutor turn and one OC response, yielding a nominal 50,000 OC responses per condition.

The baseline and RSI use the same base model within each comparison. AI agents generate the interlocutor behavior; the evaluation uses internal tests rather than production-user conversations. All response totals below pool the five models. The effective scored counts are 49,824 baseline replies and 49,793 RSI replies, and each rate uses its own denominator.

The experiment compares the existing baseline with the complete RSI framework. Model-level allocation follows the 50 × 5 × 2 design, and the outcome tables report pooled counts.

### 5.2 Operational Definition of Dialogue Errors

We adapt the Worlds, Stories, and User Preferences organization from MiniMax's Role-Play Bench [1] into the contextual diagnostic rubric in Table 2.

**Table 2. Contextual error rubric for OC-side diagnosis and annotation.**

| Dimension | Failure categories | Evidence to inspect |
|---|---|---|
| Worlds | Language or formatting defects; reference confusion; world-rule violations | Intended language, entity identities, established state, and canon |
| Stories | Repetition or stagnation; incoherent transitions; out-of-character behavior | Prior plot events, scene progression, character traits and motivations |
| User Preferences | Speaking for or ignoring the interlocutor; blocked interaction; boundary violations | Interlocutor input, agency, scene objective, and interaction boundaries |

For annotation under this rubric, a bad reply contains at least one confirmed contextual violation. A reply with several error tags contributes once to the overall bad-response count; category counts may overlap. A bare stylistic preference, an unconventional but coherent character reaction, or the absence of an explicit question is insufficient by itself. In particular, interaction failure concerns whether the response obstructs meaningful continuation in context, not whether every response ends with a question. Uncertain cases are recorded separately for review.

The annotation protocol links each confirmed label to a response, error category, and supporting context, and applies a common evaluator configuration across conditions. Section 6 reports the original binary evaluation labels; Table 2 provides the category structure for diagnostic analysis.

### 5.3 Human Evaluation and Metrics

The human evaluation contains 600 blinded pairwise judgments, each recorded as an RSI win, tie, or loss. The statistical unit in the outcome table is a judgment. We report the full distribution and the win rate conditional on a decisive judgment.

Let $E_s$ be the number of bad replies and $N_s$ the evaluated replies in condition $s$. The bad-response rate is $r_s = E_s/N_s$. The relative reduction is

$$R = 1 - r_{\mathrm{RSI}}/r_0.$$

The absolute reduction in percentage points is $100(r_0 - r_{\mathrm{RSI}})$. For human outcomes $W$, $T$, and $L$, overall win rate is $W/(W+T+L)$, decisive win rate is $W/(W+L)$, and half-credit preference is $(W+T/2)/(W+T+L)$. Half-credit preference assigns a weight of 0.5 to each tie. We report descriptive effects, with conversation and OC clustering relevant to subsequent uncertainty estimation.

## 6 Results and Analysis

### 6.1 Response Quality

The baseline produces 6,973 bad replies among 49,824 evaluated replies, compared with 2,459 among 49,793 for RSI. The bad-response rate falls from 14.00% to 4.94%. This is a reduction of 9.06 percentage points, or 64.71% relative to the baseline rate.

**Table 3. Pooled response quality across the five models.**

| Metric | Baseline | RSI |
|---|---|---|
| Evaluated OC replies | 49,824 | 49,793 |
| Bad replies | 6,973 | 2,459 |
| Bad-response rate | 14.00% | 4.94% |
| Non-bad replies | 42,851 | 47,334 |

At a common scale of 1,000 replies, the rates correspond to approximately 140 bad replies for the baseline and 49 for RSI. Each rate uses its condition-specific evaluated reply count. The complementary non-bad response rate rises from 86.00% to 95.06%.

### 6.2 Blinded Human Preference

RSI receives 294 wins, 187 ties, and 119 losses across the 600 blinded comparisons. Its overall win rate is 49.00%, while the loss rate is 19.83%. Among the 413 decisive judgments, the RSI win rate is 71.19%. The half-credit preference score is 64.58%.

**Table 4. Distribution of blinded human judgments.**

| Outcome | Count | Share |
|---|---|---|
| RSI win | 294 | 49.00% |
| Tie | 187 | 31.17% |
| RSI loss | 119 | 19.83% |
| Total | 600 | 100.00% |

Wins exceed losses by 175 judgments, or 29.17 percentage points of the complete comparison set. The tie fraction shows that many comparisons yield no clear preference, and the losses identify cases where the baseline remains preferable. The lower defect rate and the favorable balance of human wins to losses agree in direction, supporting an aggregate improvement on the internal test distribution.

### 6.3 Interpretation

The framework connects observed failures to localized changes in memory and prompting, allowing corrections to accumulate while preserving OC-specific state. The measured improvement is the combined effect of the complete system. Isolating the additional contribution of updater evolution requires a component-level comparison.

The five-model, 100-round design evaluates aggregate quality across extended conversations and several base models. Per-model differences and the distribution of gains across dialogue turns remain outside the scope of the pooled analysis.

## 7 Limitations

AI-generated interlocutors may concentrate tests around the simulator's style and scenario preferences. The findings concern controlled internal dialogue quality; transfer to production interaction requires separate evaluation. Pooled outcomes can mask variation across models, OCs, and conversations.

Uncertainty estimates require conversation-level records, repeated-judgment information, and evaluator agreement. Repeated validation feedback can also overfit selection scenarios, motivating a final test set outside the optimization loop.

## 8 Conclusion

We present a bounded RSI framework for long-horizon role-playing dialogue, with explicit interfaces for character memory, stable instructions, dynamic prompting, and updater evolution. The updater is specified through diagnostic, routing, and selection rules that can be revised from repair outcomes under fixed permissions and validation controls. In the internal five-model evaluation, the bad-response rate decreases from 14.00% to 4.94%, a 64.71% relative reduction. RSI wins 294 of 600 blinded comparisons, with 187 ties and 119 losses, giving a 71.19% win rate among decisive judgments. These results support improved aggregate dialogue quality for the complete framework across the evaluated OC conversations.